\documentclass[conference]{IEEEtran}

\IEEEoverridecommandlockouts
\usepackage{cite}
\usepackage{amsmath,amssymb,amsfonts}
\usepackage{algorithmic}
\usepackage{graphicx}
\usepackage{textcomp}
\usepackage{color,soul}
\usepackage{diagbox} % table
\usepackage{multirow}
\usepackage[table,xcdraw]{xcolor}
\def\BibTeX{{\rm B\kern-.05em{\sc i\kern-.025em b}\kern-.08em
    T\kern-.1667em\lower.7ex\hbox{E}\kern-.125emX}}

\begin{document}

\title{Efficient Fine-Tuning of BERT Models on the Edge\

\thanks{* Equal contribution.}
}

\author{
    \IEEEauthorblockN{Danilo Vucetic*,
    Mohammadreza Tayaranian*, 
    Maryam Ziaeefard,\\
    James J. Clark,
    Brett H. Meyer and
    Warren J. Gross}
\IEEEauthorblockA{\textit{Department of Electrical and Computer Engineering},\\
\textit{McGill University}\\
Montreal, Canada\\
\{danilo.vucetic, mohammadreza.tayaranian\}@mail.mcgill.ca, \\
\{maryam.ziaeefard, james.clark1, brett.meyer, warren.gross\}@mcgill.ca}} %\IEEEauthorrefmark{1} for asterisks

\IEEEoverridecommandlockouts
\IEEEpubid{\makebox[\columnwidth]{978-1-7281-0397-6/22/\$31.00~\copyright2022 IEEE \hfill} \hspace{\columnsep}\makebox[\columnwidth]{ }}
\maketitle
\IEEEpubidadjcol

\begin{abstract}
Resource-constrained devices are increasingly the deployment targets of machine learning applications. Static models, however, do not always suffice for dynamic environments. On-device training of models allows for quick adaptability to new scenarios. With the increasing size of deep neural networks, as noted with the likes of BERT and other natural language processing models, comes increased resource requirements, namely memory, computation, energy, and time. Furthermore, training is far more resource intensive than inference. Resource-constrained on-device learning is thus doubly difficult, especially with large BERT-like models. By reducing the memory usage of fine-tuning, pre-trained BERT models can become efficient enough to fine-tune on resource-constrained devices. We propose Freeze And Reconfigure (FAR), a memory-efficient training regime for BERT-like models that reduces the memory usage of activation maps during fine-tuning by avoiding unnecessary parameter updates. FAR reduces fine-tuning time on the DistilBERT model and CoLA dataset by 30$\%$, and time spent on memory operations by 47$\%$. More broadly, reductions in metric performance on the GLUE and SQuAD datasets are around $1\%$ on average.
\end{abstract}

\begin{IEEEkeywords}
Transformers, BERT, DistilBERT, NLP, Language Models, Efficient Transfer Learning, Efficient Fine-Tuning, Memory Efficiency, Time Efficiency, Edge Machine Learning
\end{IEEEkeywords}

\section{Introduction}

% Example structure for intro:
%   - Deep learning has revolutionized many machine learning tasks including NLP.
%   - Edge applications are increasingly seeking the advents of contemporary machine learning (i.e., deep learning) to better: user experience, application capabilities, technological capabilities, etc.
%   - Deep learning models in NLP such as BERT are far too big to realistically fit on resource-constrained edge devices so model sizes must be reduced.
%   - Reducing model size is not enough, as shown by TinyTL, the training-time memory requirements are FAR higher than inference-time and the achieved reductions in model size are insignificant compared to the residual training-time memory requirements (dynamic req.)
%   - In addition, computation is linked with the number of parameters to some extent (BE VERY CAREFUL HERE! SOME ML EXPERTS HATE THIS COMPARISON AND OTHERS LIKE IT WITH A BURNING PASSION!)
%   - Furthermore, memory accesses / usage are a major energy and time burden for edge devices (cite han 2015 or something like that, also cite computer architecture since RAM accesses take longer than computation) since memory accesses take 2 orders of magnitude more energy than computations. It is therefore utilitarian (choose a better word / phrase) to trade memory for computation.

% PARAGRAPH 1 GOALS: A brief overview of the issues
Large language models employing attention-based architectures have revolutionized Natural Language Processing (NLP) with the likes of BERT and GPT-3, achieving numerous state-of-the-art scores on well-studied tasks \cite{devlin2018bert, brown2020language}. As language models move from the cloud to resource-constrained devices (i.e., edge devices), their deployment is made difficult by constraints in available resources such as memory, computation, energy, and time \cite{energy_in_NLP, knowledge_transfer_edge, survey_on_device_ML}. Solely considering inference for edge learning is insufficient as the operating environment may be dynamic, and input data may change over time \cite{survey_on_device_ML}. Learning on edge devices allows models to continually adapt to their environments and to the requirements of their users, without revealing data to the cloud \cite{knowledge_transfer_edge, satyanarayanan2017emergence, survey_on_device_ML}. However, training requires about three times more memory operations than inference\footnote{During inference only the parameters and data are retrieved from memory (2 operations). During training, the parameters must be retrieved at least twice and stored once after being updated, activations are stored and retrieved, and data is retrieved (6 operations).} and more resource utilization generally. Tools like model compression are not enough to bridge the gap (i.e., a model 1/3 the size with similar performance is hard to achieve), and compressed models may not be training-efficient \cite{tinytl}. The problem specifically lies with the greater memory requirements of training compared to inference, since: 1) accesses to main memory are two orders of magnitude more energy intensive than computation; and 2) large models can not fit in cache, meaning that higher latency accesses to main memory are required \cite{tinytl, songhan_eie, comp_arch_hennessy_patterson_2012}.

Enabling on-device learning for large language models necessitates two conditions: 1) using a compressed model to reduce baseline memory requirements, and 2) training in a resource-efficient manner. To achieve the latter condition we propose \textit{Freeze And Reconfigure (FAR)}, a memory-efficient fine-tuning regime for BERT-like models. We focus first and foremost on reducing memory usage during the backward pass by reducing the largest contributors to said usage: activations and gradients. We introduce a methodology that identifies which parameters to \textit{freeze} and which to continue training. \textit{Freeze} here refers to disabling parameter updates during back-propagation. We also propose a method of dynamic architecture reconfiguration such that frozen and non-frozen parameters are grouped separately. Reconfiguration reduces poorly-structured memory accesses which, for example, encumber unstructured pruning methods. The contributions of this paper are summarized as follows:
\begin{itemize}
%    \item We propose FAR, a novel technique to reduce the memory footprint of fine-tuning BERT-based models.
    \item We introduce a novel learning metric that tracks the size of weight updates over contiguous groups of parameters during fine-tuning using the $L_1$ norm. According to the metric, the best-learning subsets of nodes are selected.
    \item We introduce a novel parameter freezing scheme that exploits architectural reconfiguration, grouping frozen and nonfrozen parameters, reducing training-time resource consumption.
%    \item We show that FAR reduces memory usage and latency during fine-tuning while maintaining metric performance.
\end{itemize}

FAR achieves sizeable reductions in the resource consumption of fine-tuning. Our results show that when fine-tuning DistilBERT on CoLA with 60$\%$ of the model parameters frozen, fine-tuning time is reduced by 30$\%$ and memory access time is reduced by 47$\%$, while metric performance is maintained. Similar results are achieved for other GLUE and SQuAD tasks. Evidently, FAR makes training more efficient while maintaining metric performance. 

\section{Background and Related Work}

% Example structure for background and related works:
%   - We should first explain some basics about Transformer-based models such as their architecture, the bottlenecks, the training methods etc. These will give the reader a good idea of why and how we can reduce memory usage. We should also give compressed models here and say that these are vital in decreasing memory usage during training. 
%   - We can then explain methods similar to ours such as bitfit

\subsection{Transformers and BERT Models: Training and Efficiency}
%Explain some stuff about Transformers and BERT models. Explain the training mechanisms that are used. Explain fine-tuning vs pre-training etc.

Transformers and BERT-like models make use of Attention and Feed-Forward Network (FFN) sublayers to extract linguistic knowledge and achieve state-of-the-art scores\cite{vaswani2017attention}. BERT in particular uses the Transformer Encoder architecture to model languages by pre-training on large datasets of relatively simple language tasks. To deploy BERT-like models to a new downstream task, the model must be fine-tuned \cite{devlin2018bert}. This allows the use of a single set of weights, the pre-trained BERT weights, as the starting point for various, faster converging, fine-tuning tasks, avoiding the large overhead of pre-training the full BERT model\footnote{As reported in \cite{devlin2018bert}, 4 cloud tensor processing units running for 4 days are required to pre-train BERT's 110 million parameters.}. Various methods have been proposed to compress BERT including DistilBERT \cite{distilbert}, MobileBERT \cite{sun-etal-2020-mobilebert}, and TinyBERT \cite{tiny_bert}. A detailed discussion of compressing large language models is found in \cite{Ganesh2021CompressingLT}, including analysis of the major bottleneck of Transformer-based models: the FFN sublayers, which contribute more parameters, runtime memory consumption, and inference latency than all other sublayers. Model compression is useful for enabling on-device learning, but greater efficiency during training is desirable to further enable the realization of large models on edge devices.  

%A common method to avoid training overheads is to train the model on general unlabeled data so that it gains a general understanding of the language. The pre-trained model is then used for transfer learning to fine-tune the pre-trained model's parameters for each down-stream task. BERT \cite{devlin2018bert} is a notable example of such pre-trained, attention-based language models. The pre-training is done on cloud devices while the less resource consuming fine-tuning is performed on edge devices.

\subsection{Efficient Learning}
%Bitfit and related techniques should go here. 

On-device learning has been detailed in \cite{survey_on_device_ML}, including methods of data compression and theoretical approaches to efficiency. When considering on-device learning, a fundamental trade-off must be kept in mind to train in a reasonable amount of time: either the training-time resource requirements of a model must be decreased, or the hardware resources must be increased. The former approach is usually studied in terms of model compression such as in \cite{distilbert, sun-etal-2020-mobilebert, tiny_bert}. Others take the problem as reducing the real requirements of training without altering model architecture. One such approach, \cite{weight_update_pruning}, trains only the best-learning weights of a feed-forward neural network, while keeping the others at their initialization values. Trained weights are selected by tracking the largest accumulated weight updates over some initial training steps. The authors also further reduce memory requirements by regenerating frozen weights from a pseudorandom number generator, thereby trading memory for extra computation. This approach does not work for BERT-like models as the weights of the pre-trained model are not randomly generated. Additionally, the unstructured selection of weights leads to inefficient memory accesses, an issue which we address with structured selection in FAR. A similar approach, BitFit \cite{bitfit}, freezes all BERT weights during fine-tuning and only applies updates to bias terms. The authors report near-baseline performance despite the large drop in trained parameters. While BitFit works on uncompressed (i.e., highly overparameterized) BERT models, on the more compressed variants like DistilBERT, it is expected to fail due to the smaller network size and lower capacity. We address this issue by fine-tuning a greater proportion of the parameters. 

\section{Freeze And Reconfigure} \label{method_far}

On-device learning of large language models requires compressed models and efficient training procedures. The reduction in memory usage provided by these requirements has knock-on effects: higher energy efficiency, reduced computation, and reduced memory access times. In the case of memory access time, reducing the number of parameters decreases the amount of information transferred across a system. 

FAR is specifically designed for use with compressed language models, specifically DistilBERT, whose capacity and adaptability are diminished from their uncompressed counterparts \cite{mirzadeh2020improved}. This is accomplished by fine-tuning a subset of linear layer weights (called FFNs in DistilBERT) as well as their biases, in contradistinction to BitFit, which trains only the biases of BERT. 

In FAR, the sets of fine-tuned weights in the linear layers are selected in a structured manner to avoid sparse memory accesses which would delay memory operations \cite{wen2016learning}. Each node in an FFN layer is considered a single group of parameters. These nodes are classified as \textit{learner} and \textit{nonlearner} nodes depending on how well a node performs during fine-tuning. The parameters of the nonlearner nodes are frozen while learner nodes are fine-tuned. Frozen nodes no longer require gradient calculations during the backward pass, storage of activations, or memory accesses during the backward pass. Reconfiguration of the FFN layers is completed during fine-tuning by separating learner and nonlearner nodes into distinct, newly created, FFN sublayers.

%{and has benefits in the cache of potentially higher hit rates.%} 

%The idea behind this approach is to utilize the stored information in nonlearner weights from pre-training and allowing the learner nodes to learn the down-stream task. To help with the learning, all the bias terms are also updated.

%Each node in the FFN layers is evaluated on a layer-by-layer basis to determine if it is an effective learner

%The proposed method aims at reducing the memory access time of fine-tuning BERT-based models by restricting the set of trained parameters. FAR divides the FFN nodes into two groups: \textit{Learner} and \textit{nonlearner} nodes. The weights of the former node group are normally updated while the latter are frozen. 

%\begin{figure}[tbp]
%\centerline{\includegraphics[width=0.45\textwidth]{images/distilbert.png}}
%\caption{Overview of a Transformer block in Distilbert. The FFN consists of two dense layers with the matrix sizes shown in the picture.}
%\label{fig:distilbert}
%\end{figure}

Each Transformer block in DistilBERT is made up of a multi-head attention layer which feeds an FFN. The FFN consists of two dense layers which contain 3072 and 768 nodes respectively. The weights of the FFNs make up more than $66\%$ of the parameters of the whole network. Thus, freezing a subset of these weights has a considerable effect on the overall number of updated parameters of the model. Past work has shown that freezing up to 80$\%$ of feed-forward neural network parameters results in minimal changes in accuracy for smaller models and simpler tasks \cite{weight_update_pruning}. %Therefore, it is expected that the FFNs of DistilBERT can adapt to a reduction in trained parameters of at least 50$\%$ while fine-tuning with FAR.

\subsection{Selection of Learner Nodes} \label{selection}

The set of learner nodes is decided upon after an initial set of fine-tuning iterations during a process we call \textit{priming} (cf. Figure \ref{fig:priming}, green nodes in left figure are learner nodes). During priming, a copy of the pre-trained FFN weights is stored and the network is fine-tuned for some percentage, $p$, of the total number of optimization steps. The priming percentage, $p$, is kept small so as to avoid the additional burden of fully fine-tuning a large model for a greater proportion of the optimization steps. It is expected, however, that more priming should result in higher metric performance. After fine-tuning the model during priming, the initial FFN weight vectors, $\boldsymbol{w}^{e, i}_{n}$, are subtracted from the fine-tuned FFN weight vectors, $\boldsymbol{\phi}^{e, i}_{n}$, for each Transformer Encoder, $e$, FFN sublayer, $i$, and node, $n$. We employ the $\text{L}_1$ norm to compute the total amount learned by each node in the FFN. This is in distinction to \cite{weight_update_pruning}, where each weight is chosen individually based on the difference in its initial and trained value. Usage of the $\text{L}_1$ norm allows for the weights of the FFN node to be grouped so that architectural modifications can be made later for greater memory savings. The learning metric is computed as: 

\begin{equation}
    m^{e, i}_{n} = \left \|\boldsymbol{\phi}^{e, i}_{n} - \boldsymbol{w}^{e, i}_{n}  \right \|_1
    \label{eq:l1}
\end{equation}
% $e \in \left \{ 1, \cdots, 6 \right \}$
% $i \in \left \{ 1, 2 \right \}$
% $n \in \left \{ 1, \cdots, l_i \right \}$ where $l_i$ is the number of nodes in FFN sublayer

Finally, to compute the set of learner nodes, a retention percentage, $r$, is defined as the ratio of learner nodes to the total number of nodes. After sorting the learning metric for each FFN, the $r\%$ of nodes with the largest metric are classified as learners; the rest as nonlearners. Note that decreasing the node retention results in more nonlearner nodes, which in turn reduces memory utilization during fine-tuning. Using higher retention values is expected to have a positive impact on metric performance as it increases the model's ability to adapt to the down-stream task. In addition, the priming and retention hyperparameters are defined as percentages to allow them to scale to larger datasets and varying models. For example, a more compressed model may require higher node retention to avoid substantial losses in metric performance. 

\subsection{Dynamic FFN Reconfiguration} \label{reconfig}

\begin{figure}[tbp]
\centerline{\includegraphics[width=0.45\textwidth]{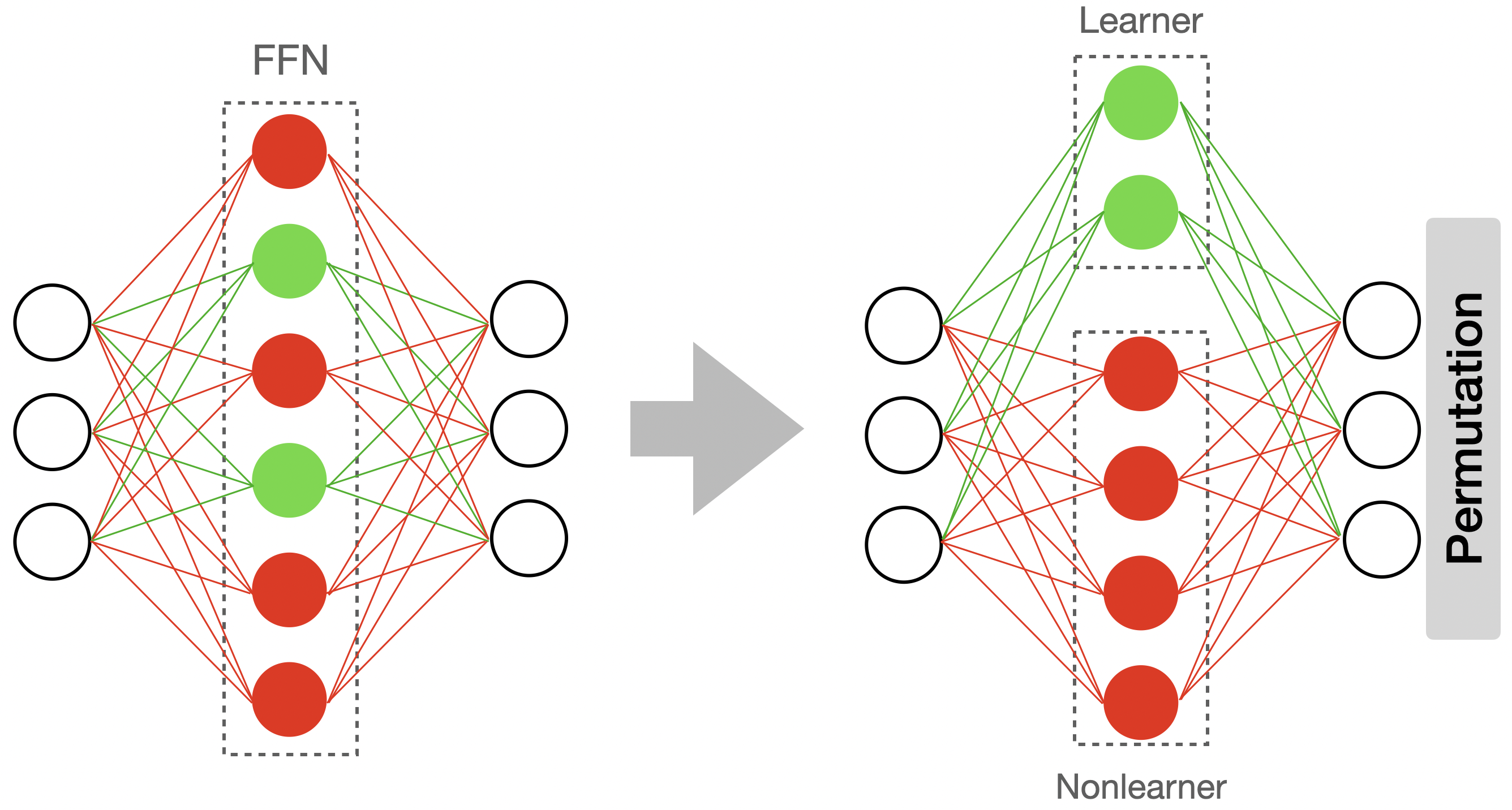}}
\caption{Overview of priming steps and the dynamic reconfiguration of nodes in FFN. In the left, the learner (green) and nonlearner (red) nodes are chosen after the priming steps. In the right, the nodes are reconfigured to disable gradient computation for red weights. The output of the FFN is permuted after reconfiguration.}
\label{fig:priming}
\end{figure}

After assigning the learner and nonlearner nodes, the FFNs of each Transformer block are reconfigured to separate the nonlearner nodes from learner nodes. As shown in Figure \ref{fig:priming}, FFN nodes are reconfigured into two parallel sub-modules. Such reconfiguration ensures that the memory of learner nodes is grouped together, allowing only the fine-tuned parameters to be retrieved during fine-tuning. Finally, the reconfigured FFNs need to have their output order restored such that their outputs are coherent with the parameters of the model, so a permutation is applied to the output vectors. The combination of reconfiguration and permutation trades memory accesses for additional time spent on computation and data movement, which is far less costly \cite{songhan_eie}.

Using nodes as parameter groupings and reconfiguring the model this way allows us to avoid memory access issues that arise in methods like unstructured pruning. Since individual weights are never solely accessed, we avoid sparse memory accesses. Reconfiguration is also beneficial for practical implementation since in PyTorch, automatic gradient computation can not be disabled for singular parameters, but instead only for whole layers or sublayers. Thus, unlike unstructured pruning, no special hardware or libraries are needed to implement FAR.

%implementation used PyTorch, where automatic gradient computation can only be disabled for a group of parameters, thus it is convenient to reconfigure the FFN and freeze just the nonlearner nodes. 

% In addition to the benefits of reducing memory operations, this can lead to better cache performance since no frozen parameters are retrieved from main memory, meaning that fewer cache misses should be expected.

\section{Experiments}

\begin{table}
\centering
\caption{Effectiveness of priming and retention with 5 averaged runs on DistilBERT for each dataset}
\begin{tabular}{|c|ccc|ccc|}
\cline{2-7}
\multicolumn{1}{l|}{} & \multicolumn{3}{c|}{MNLI}                                                & \multicolumn{3}{c|}{CoLA}                                                \\ \hline
\diagbox{$p$}{$r$}                    & \multicolumn{1}{c|}{10$\%$}   & \multicolumn{1}{c|}{25$\%$}  & 40$\%$            & \multicolumn{1}{c|}{10$\%$}   & \multicolumn{1}{c|}{25$\%$}  & 40$\%$            \\ \hline
1$\%$                   & \multicolumn{1}{c|}{81.8} & \multicolumn{1}{c|}{81.9} & 82.1          & \multicolumn{1}{c|}{52.8} & \multicolumn{1}{c|}{53.1} & 53.3          \\ \hline
5$\%$                   & \multicolumn{1}{c|}{81.8} & \multicolumn{1}{c|}{82.1} & 81.9          & \multicolumn{1}{c|}{51.9} & \multicolumn{1}{c|}{52.7} & 53.7          \\ \hline
10$\%$                    & \multicolumn{1}{c|}{81.9} & \multicolumn{1}{c|}{82.0} & \textbf{82.2} & \multicolumn{1}{c|}{53.4} & \multicolumn{1}{c|}{52.5} & \textbf{54.0} \\ \hline
\end{tabular}
\label{tab:grid_search}
\end{table}

% mention dataset sizes
\begin{table*}
\centering
\caption{Metric performance of DistilBERT with and without FAR on GLUE and SQuAD 2.0 - $1\%$ priming and $10\%$ and $40\%$ retention. Each FAR run is accompanied with its percentage drop compared to the baseline. The last column shows the average score drop percentage across all tasks. $\dagger$ Shows our implementation.}
\begin{tabular}{c|c|c|c|c|c|cc|c|}
\cline{2-9}
\textbf{}                                           & CoLA           & MNLI             & QNLI           & QQP              & SST2            & \multicolumn{2}{c|}{SQuAD 2.0}         &                                            \\
                                                    & Matthews Corr. & Acc.             & Acc.           & Acc.             & Acc.            & \multicolumn{1}{c|}{EM}      & F1      & \multirow{-2}{*}{Avg}                      \\ \hline
\multicolumn{1}{|c|}{Baseline $\dagger$}                & \textbf{53.6} & \textbf{82.1}   & \textbf{89.1} & \textbf{90.2}   & 91.1           & \multicolumn{1}{c|}{\textbf{65.0}}  & \textbf{67.9}  & \cellcolor[HTML]{FFFFFF}                   \\ \cline{1-8}
\multicolumn{1}{|c|}{}                              & 52.8          & 81.9            & 88.2          & 90.0              & 91.1           & \multicolumn{1}{c|}{63.3}  & 66.4  & \multirow{-2}{*}{\cellcolor[HTML]{FFFFFF}} \\ \cline{2-9} 
\multicolumn{1}{|c|}{\multirow{-2}{*}{FAR$_{10}$}} & -1.58\%        & -0.24\% & -1.01\%        & -0.22\% & 0.00\% & \multicolumn{1}{c|}{-2.59\%} & -2.27\% & -1.13\%                                    \\ \hline
\multicolumn{1}{|c|}{}                              & 53.3          & \textbf{82.1}   & 88.5          & \textbf{90.2}   & \textbf{91.3}  & \multicolumn{1}{c|}{63.6}  & 66.6  & \cellcolor[HTML]{FFFFFF}                   \\ \cline{2-9} 
\multicolumn{1}{|c|}{\multirow{-2}{*}{FAR$_{40}$}} & -0.55\%        & 0.00\%           & -0.67\%        & 0.00\%           & 0.22\%          & \multicolumn{1}{c|}{-2.14\%} & -2.01\% & -0.74\%                                    \\ \hline
\end{tabular}
\label{tab:main_results}
\end{table*}

% Things to mention:
% - the model we're using and a little explanation why: DistilBERT. Also say we haven't done a hyperparameter search and give the specific learning rate and number of epochs. 
% - the datasets we're using and why: some GLUE, and SQuAD 2.0
% - Mention that we are running everything on V100 to get the accuracy results and we run on Jetson to get the timing and memory time results. 
% - what experiments we did and why: 
%       - grid search of hyper-parameters to find a good accuracy-resource trade-off. Mention why we choose 1% priming, and however many % retention. basically it provides good results. use the figures!
%       - using results from above, we chose two setups with different goals FAR_10 and FAR_40, show the results of these and give the metric drop (include this all in a table.) 
%       - Give the results for timing -> our method has lower memory access time and is faster to train. Shows the trade-off in action!
%       - Ablation study -> show that we do better than random selection and we do better than BitFit, which is important because it means our method works for compressed models!

We test FAR using the compressed BERT-based model DistilBERT \cite{distilbert}. DistilBERT is a pre-trained language model and can be fine-tuned to downstream tasks. DistilBERT has six Transformer encoder layers, each with two FFN sublayers (i.e., $e \in \left \{ 1, \cdots, 6 \right \}$ and $i \in \left \{ 1, 2 \right \}$ in Equation \ref{eq:l1}). We fine-tune the model using a learning rate of $2e^{-5}$, with a linear learning rate schedule, batch size of 16, using at most 5 epochs.

Experiments are run on \textit{NVIDIA Tesla V100 Volta} GPU accelerators to gather metric results for all datasets. Wherever timing is reported, these results have been gathered on the \textit{NVIDIA Jetson Xavier NX} edge device. Memory access time is computed using the NVIDIA's profiling tool, \textit{nvprof}. Memory access time is the summation of the execution time for memory-related API calls, namely \textit{cudaStreamSynchronize, cudaMemcpyAsync} and \textit{cudaMemsetAsync}.

% explain what these API calls do.

FAR is tested against conventional and efficient fine-tuning approaches using standard NLP datasets from the GLUE benchmark and SQuAD 2.0 \cite{wang2018glue, squad2}. All the GLUE tasks are included in the experiments except for RTE, MRPC, STS-B because of their small dataset sizes and WNLI because of its confounding results \cite{gordon2020compressing}. This leaves MNLI, QQP, QNLI, SST-2, and CoLA. CoLA (9594 training examples) and MNLI (433k training examples) are used to find effective priming and retention percentages in grid searches (Cf. Table \ref{tab:grid_search}). MNLI and CoLA are believed to represent small and large datasets, respectively \cite{wang2018glue}. Each reported value is the average of 5 fine-tuning attempts with random seeds to ensure consistency. 

In order to test the hypothesis that higher priming and higher retention percentages produce better metric performance, we complete a grid-search style test on CoLA and MNLI. The results are listed in Table \ref{tab:grid_search}. The table shows that with higher retention percentage and higher priming percentage, fine-tuning produces a higher metric score, with the highest score occurring with $10\%$ priming and $40\%$ retention. Moreover, the difference between $1\%$ and $10\%$ priming is small enough to warrant the use of $1\%$ priming for FAR. This also means that the training overhead is made minimal since extra priming does not necessarily correlate to better performance. Henceforth, $\text{FAR}_x$ will represent FAR using $1\%$ priming and $x\%$ node retention. $\text{FAR}_{10}$ and $\text{FAR}_{40}$ are used to demonstrate two trade-off points between metric performance and resource utilization. The former has lower resource utilization while also producing lower metric performance and vice-versa for the latter. The results of this comparison are listed in Table \ref{tab:main_results} alongside with the baseline implementation of DistilBERT using the Hugging Face python library\cite{huggingface}, without freezing any weights during fine-tuning. We observe higher metric scores for $\text{FAR}_{40}$ compared to $\text{FAR}_{10}$ and a lower drop in performance compared to the baseline at just $0.74\%$ compared to $1.13\%$ respectively. The largest drops in performance occur on SQuAD 2.0, which is the most complex dataset tested.
%ADD IF WE HAVE SPACE
%The drop in accuracy is expected as FAR uses fewer parameters for training, however, the relative drop in the performance is very low compared to the baseline.

\begin{figure}[tbp]
\centerline{\includegraphics[width=0.45\textwidth]{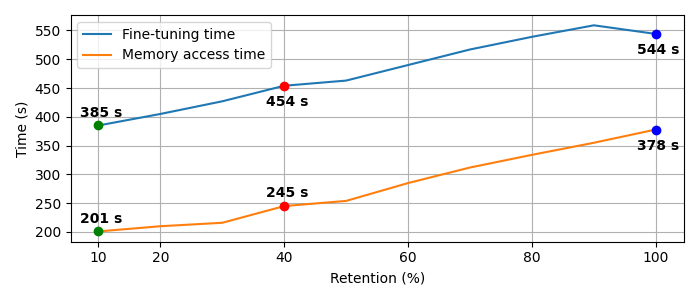}}
\caption{Overall time and total memory access time for fine-tuning DistilBERT on CoLA using FAR with various retention percentages on  the NVIDIA  Jetson  Xavier  NX. Priming percentage is set to 1\% across all runs. Values for FAR$_{10}$, FAR$_{40}$ and the baseline (FAR$_{100}$) are indicated with green, red and blue dots, respectively.}
\label{fig:jetson}
\end{figure}

Figure \ref{fig:jetson} shows the effect of changing the retention percentage on training time and memory access time. A similar trend is observed in both figures, showing that using setups with lower retention values reduces both fine-tuning and memory access time. These results indicate that FAR reduces the resource consumption of fine-tuning, and as a result also reduces training time. $\text{FAR}_{10}$ and $\text{FAR}_{40}$ are also indicated with green and red dots, respectively. In terms of overall fine-tuning time, $\text{FAR}_{10}$ and $\text{FAR}_{40}$ each respectively show $30\%$ and $17\%$ reductions compared to the baseline. For memory access time, $47\%$ and $36\%$ are the time reductions associated to $\text{FAR}_{10}$ and $\text{FAR}_{40}$.

\begin{table}
\centering
\caption{Comparison of FAR$_{10}$ with random node selection and BitFit}
\begin{tabular}{c|c|c|cc|}
\cline{2-5}
\multirow{2}{*}{\textbf{}}                                                            & CoLA          & MNLI          & \multicolumn{2}{c|}{SQuAD 2.0}                     \\ \cline{4-5} 
                                                                                      & Mathews Corr. & Acc.          & \multicolumn{1}{c|}{EM}            & F1            \\ \hline
\multicolumn{1}{|c|}{FAR$_{10}$}                                                     & \textbf{52.8} & \textbf{81.9} & \multicolumn{1}{c|}{\textbf{63.3}} & \textbf{66.4} \\ \hline
\multicolumn{1}{|c|}{\begin{tabular}[c]{@{}c@{}}Random\\ Selection$_{10}$\end{tabular}} & 52.0          & 81.8          & \multicolumn{1}{c|}{62.8}          & 65.7          \\ \hline
\multicolumn{1}{|c|}{BitFit}                                                          & 51.7          & 81.5          & \multicolumn{1}{c|}{49.8} & \multicolumn{1}{c|}{49.8}        \\ \hline
\end{tabular}
\label{tab:ablation}
\end{table}

Finally, we carry out an ablation study to measure the effectiveness of our method against related works and to show that FAR is in fact necessary for efficient fine-tuning. First, to show the effect of using priming, we perform an experiment in which the learner and nonlearner nodes are randomly selected. Table \ref{tab:ablation} shows that for the same retention value of $10\%$, FAR, which uses priming to select nodes, performs better than random node selection, with the gap being wider in the more complex SQuAD task. Comparing FAR with BitFit, which freezes all the weights of all dense layers across the network, we see a drop in performance of less than 2$\%$ for CoLA and MNLI. However, BitFit performs poorly on SQuAD 2.0, proving its failure to cope with complex tasks on compressed models like DistilBERT. The memory access and training times of fine-tuning on DistilBERT using BitFit on CoLA are 237 and 384 seconds. Clearly, BitFit spends more time accessing memory, but around the same time fine-tuning as FAR$_{10}$ (Figure \ref{fig:jetson}).

\section{Conclusions and Future Work}
In this paper we propose FAR, a novel method to reduce fine-tuning memory consumption by exploiting the overparameterization of large language models. We show that for a compressed BERT-based model, it is not enough to simply freeze large sections of the model during fine-tuning, but that freezing must be selective and structured. Despite large decreases in the number of fine-tuned parameters, metric performance remains near baseline levels, while fine-tuning time and memory usage are significantly reduced. Future work involves investigations into new domains such as Transformers on NLP and vision tasks. We will also apply FAR to other backbone models such as MobileBERT to demonstrate the generalizability of the method. 

\section*{Acknowledgment}
We thank Huawei Canada for their sponsorship of this work and Compute Canada for providing computational resources.

%Additionally, novel mechanisms for learner node selection must be investigated to reduce overhead and resource consumption while increasing metric performance. Further architectural modifications will also be investigated to increase performance. These may include the addition of extra small layers to supplement lost capacity, the replacement of sublayers such as self-attention to improve computational complexity, and pre-training of novel efficient model architectures.

\bibliographystyle{IEEEtran}
\bibliography{references}

\end{document}